\documentclass[referee,sn-vancouver]{sn-jnl}

\renewenvironment{table}[1][]%
{\tableorg[#1]%
\tablebodyfont%
\renewcommand\footnotetext[2][]{{\removelastskip\vskip3pt%
\let\tablebodyfont\tablefootnotefont%
\hskip0pt\if!##1!\else{\smash{$^{##1}$}}\fi##2\par}}%
}{\endtableorg}
\usepackage[utf8]{inputenc}
\usepackage{graphicx}
\usepackage[ruled]{algorithm2e}
\usepackage{hyperref}
\usepackage{verbatim}
\usepackage{amsmath,amssymb,amsfonts}
\usepackage{amsthm}
\usepackage{mathrsfs}
\usepackage[title]{appendix}
\usepackage{xcolor}
\usepackage{textcomp}
\usepackage{manyfoot}
\usepackage{tabularray}
\UseTblrLibrary{booktabs}
\usepackage{listings}
\usepackage{rotating}
\usepackage{array}
\usepackage{multirow}
\usepackage{acro}
\usepackage{tabularx}
\usepackage{float}
\usepackage{pdflscape}
\usepackage{svg}
\usepackage{natbib}
\usepackage{orcidlink}
\usepackage{float}
\usepackage{fvextra}   
\usepackage{chngcntr}  

\newcounter{promptgroup}
\newcounter{prompt}[promptgroup]

\renewcommand{\theprompt}{\thepromptgroup.\arabic{prompt}}

\newenvironment{promptgroupenv}[1][]{%
  \refstepcounter{promptgroup}%
  \begin{quote}
    \noindent\rule{\linewidth}{0.6pt}
    \textbf{Prompt Group}%
    \ifx\relax#1\relax\else: \textit{#1}\fi
    
    \noindent\rule{\linewidth}{0.6pt}
}{%
  \end{quote}
}

\newenvironment{prompt}[1][]{%
  \refstepcounter{prompt}%
  \begin{quote}
    \textbf{Prompt \theprompt}%
    \ifx\relax#1\relax\else\ \textit{#1}\fi
    \par\smallskip
}{%
  \end{quote}
}

\DefineVerbatimEnvironment{prompttext}{Verbatim}{
  breaklines=true,
  breakanywhere=true
}

\newcommand{\promptgroupcaption}[1]{%
  \par\smallskip
  \noindent
  \begin{center} 
  \textbf{Pro. \thepromptgroup} #1
  \end{center}
  \smallskip
}

\newcommand{\fref}[1]{\footnote{\href{#1}{#1}}}

\title{Automatic Cardiac Risk Management Classification using large-context Electronic Patients Health Records}

\author[1]{\fnm{Jacopo} \sur{Vitale}}\email{jacopo.vitale@unicampus.it}
\author[1]{\fnm{David} \sur{Della Morte}}
\author[1]{\fnm{Luca} \sur{Bacco}}
\author[1]{\fnm{Mario} \sur{Merone}}
\author[2]{\fnm{Mark C.H.} \sur{de Groot}}
\author[2]{\fnm{Saskia} \sur{Haitjema}}
\author[1]{\fnm{Leandro} \sur{Pecchia}}
\author[2]{\fnm{Bram} \spfx{van} \sur{Es}\email{bes3@umcutrecht.nl}}

\affil[1]{\orgdiv{Research Unit of Intelligent Health Technologies}, \orgname{Università Campus Bio-Medico di Roma}, \orgaddress{\city{Rome}, \country{Italy}}}
\affil[2]{\orgdiv{Central Diagnostic Laboratory}, \orgname{University Medical Center Utrecht}, \orgaddress{\city{Utrecht}, \country{The Netherlands}}}

\DeclareAcronym{nlp}{
  short=NLP,
  long=natural language processing,
}
\DeclareAcronym{ehr}{
  short=EHR,
  long=electronic health record,
}
\DeclareAcronym{cvrm}{
  short=CVRM,
  long=Cardiovascular Risk Management,
}
\DeclareAcronym{lhs}{
  short=LHS,
  long=Learning Healthcare System,
}
\DeclareAcronym{ner}{
  short=NER,
  long=named entity recognition,
}
\DeclareAcronym{all}{
  short=ALL,
  long=approximate list lookup,
}
\DeclareAcronym{cnn}{
  short=CNN,
  long=convolutional neural network,
}
\DeclareAcronym{umcu}{
  short=UMCU,
  long=University Medical Center of Utrecht,
}
\DeclareAcronym{bow}{
  short=BOW,
  long=bag-of-words,
}
\DeclareAcronym{icd}{
  short=ICD,
  long=International Classification of Disease,
}
\DeclareAcronym{crf}{
  short=CRF,
  long=conditional random fields,
}
\DeclareAcronym{rnn}{
  short=RNN,
  long=recurrent neural network,
}
\DeclareAcronym{svm}{
  short=SVM,
  long=support vector machine,
}
\DeclareAcronym{lstm}{
  short=LSTM,
  long=long short-term memory,
}
\DeclareAcronym{bilstm}{
  short=biLSTM,
  long=bilateral LSTM,
}
\DeclareAcronym{tf-idf}{
  short=TF-IDF,
  long=term frequency-inverse document frequency,
}
\DeclareAcronym{gru}{
  short=GRU,
  long=gated recurrent unit,
}
\DeclareAcronym{qrnn}{
  short=QRNN,
  long=quasi-recurrent neural network,
}
\DeclareAcronym{bigru}{
  short=biGRU,
  long=bidirectional GRU,
}
\DeclareAcronym{cvd}{
  short=CVD,
  long=cardiovascular disease,
}
\DeclareAcronym{cvr}{
  short=CVR,
  long=cardiovascular risk,
}
\DeclareAcronym{vit}{
  short=ViT,
  long=Vision Transformer,
}
\DeclareAcronym{llm}{
    short = LLM,
    long = Large Language Model
}

\begin{document}
\abstract{
To overcome the limitations of manual administrative coding in geriatric Cardiovascular Risk Management, this study introduces an automated classification framework leveraging unstructured Electronic Health Records (EHRs). Using a dataset of 3,482 patients, we benchmarked three distinct modeling paradigms on longitudinal Dutch clinical narratives: classical machine learning baselines, specialized deep learning architectures optimized for large-context sequences, and general-purpose generative Large Language Models (LLMs) in a zero-shot setting. Additionally, we evaluated a late fusion strategy to integrate unstructured text with structured medication embeddings and anthropometric data. Our analysis reveals that the custom Transformer architecture outperforms both traditional methods and generative \acs{llm}s, achieving the highest F1-scores and Matthews Correlation Coefficients. These findings underscore the critical role of specialized hierarchical attention mechanisms in capturing long-range dependencies within medical texts, presenting a robust, automated alternative to manual workflows for clinical risk stratification.
}

\maketitle

\section{Introduction}\label{Intro}
In the geriatric population, cardiovascular risk represents a significant challenge due to the increased prevalence of cardiovascular disease (CVD) with advancing
age~\cite{who,benjamin}.  Older adults often exhibit a complex clinical profile, where multiple factors
including comorbidities, age-related physiological changes, frailty, and polypharmacy
synergistically contribute to an elevated cardiovascular risk~\cite{lakatta}. Although clinical guidelines exist to support healthcare professionals in managing cardiovascular risk, they are not always consistently implemented in routine practice and may become outdated as new evidence emerges~\cite{Giezeman2017, Dixit2021}. This results in suboptimal screening and management of cardiovascular risk factors, ultimately contributing to avoidable morbidity and mortality.

A Learning Healthcare System (\acs{lhs}) provides a promising framework to address these limitations by continuously analyzing data from routine clinical care, identifying gaps and inefficiencies, and 
translating new evidence into actionable improvements in practice. 
First introduced in 2007, the \acs{lhs} was conceptualized as a system in which care delivery and 
evidence generation are tightly integrated in a continuous cycle of learning and improvement. 

Formally, an \acs{lhs} is defined as a healthcare system that systematically learns from clinical 
experience by generating and applying evidence to support medical decision-making, foster innovation, 
and enhance the quality, safety, and efficiency of care delivery~\cite{Laurijssen2024,Tsai2025}.
Within the \acs{lhs} framework, Electronic Health Records (\acs{ehr}s) serve as the fundamental infrastructure that enables the transformation of routine clinical practice into systematic knowledge. \acs{ehr}s function as a centralized access point for patient data and include both structured information, 
such as laboratory values, and unstructured content, such as prescribed medications, medical notes, and information related to appointments and hospitalizations. By leveraging the data routinely captured in \acs{ehr}s, it becomes possible to automatically accumulate longitudinal patient information without the need for dedicated follow-up. This results in the creation of a dynamic and scalable database that supports predictive modeling and real-time risk stratification.
Furthermore, the integration of \acs{ehr} data facilitates the early identification and continuous monitoring of high-risk individuals, thereby enabling the implementation of personalized preventive and therapeutic strategies~\cite{hersh,Katehakis2006}.
In 2016, the Center for Circulatory Health at the University Medical Center Utrecht (UMCU), 
in the Netherlands, launched the Utrecht Cardiovascular Cohort – Cardiovascular Risk Management 
(UCC-CVRM) initiative, which constitutes one of the first operationalizations of a cardiovascular \acs{lhs}.
Following the guidelines set by the Utrecht Cardiovascular Cohort, the cardiovascular \ac{lhs} aims to uniformly assess and record risk indicators for all patients included in the \ac{cvrm}~\cite{Asselbergs2017}. 
This approach allows data generated from routine check-ups to be easily analyzed, enabling continuous improvement in the delivery of healthcare services~\cite{Asselbergs2017}.

\noindent At \acs{umcu} patients are considered eligible to \acs{cvrm} via codes in their \ac{ehr} namely, \textit{agenda codes}, that denote specific types of appointments, consultations, or healthcare services.
This eligibility assignment system has significant limitations, including the fact that it does not account for the patient's physiological parameters or clinical history, as these codes can change over time and may not accurately reflect the patients’ clinical history. 
Moreover, the assigment code is assigned manually by a nurse leading to selection errors.

This study leverages the extensive \acs{ehr} database of the \ac{umcu}, 
which includes a wide range of patient information such as demographics, medical history, diagnoses, 
treatments, and laboratory results, representing a valuable resource for enhancing \acs{cvrm}. 
Specifically, the contributions of this work are threefold: (i) To collect and preprocess highly unstructured data from the UMCU \acs{ehr} system, 
    transforming it into a structured dataset suitable for downstream analysis; (ii) To develop an automated strategy for assessing patient eligibility for the \acs{cvrm} program, with particular emphasis on the integration of clinical history data; (iii)  To explore and evaluate various model architectures and techniques capable of handling 
    long-context information, in order to improve the identification of high-risk patients.

\section{Dataset}
From \acs{umcu} database we retrieved all record from $3482$ patients \acs{ehr}s collected from the geriatric outpatient clinic. 
Each patient's medical record is assigned an identification code based on the type of appointment or visit, without considering any physiological parameters. Among the twenty-six possible codes that can be assigned, four designate the automatic enrollement to the CVR management programme and identified as eligible.   

A detailed summary of the dataset characteristics is presented in Table~\ref{tab:dataset_summary}.
Patients were categorized into two classes based on their eligibility for the \acs{cvrm} program: 
\textbf{class 0} denotes \textit{non-eligible} patients, while \textbf{class 1} denotes \textit{eligible} patients. As shown in Table~\ref{tab:dataset_summary}, the dataset is strongly imbalanced with respect to the class distribution.

\begin{table}[H]
    \centering
    \caption{Demographic and class distribution of the dataset.}
    \label{tab:dataset_summary}
    \begin{tabular}{lc}
        \hline
        \textbf{Characteristic}               & \textbf{Value}              \\
        \hline
        Male samples                          & 1871 (53.73\%)             \\
        Female samples                        & 1611 (46.27\%)             \\
        Age (mean ± SD)                       & M: 74.52 ± 9.05, F: 74.40 ± 9.51 \\
        Class 0                               & 2808 (80.61\%)             \\
        Class 1                               &   674 (19.39\%)             \\
        \hline
        \textbf{Total number of samples}      & \textbf{3482}              \\
        \hline
    \end{tabular}
\end{table}

Each patient has an undetermined number of medical text or medications based on his/her own medical history.

\begin{figure}[H]
    \centering
    \includegraphics[width=0.8\linewidth]{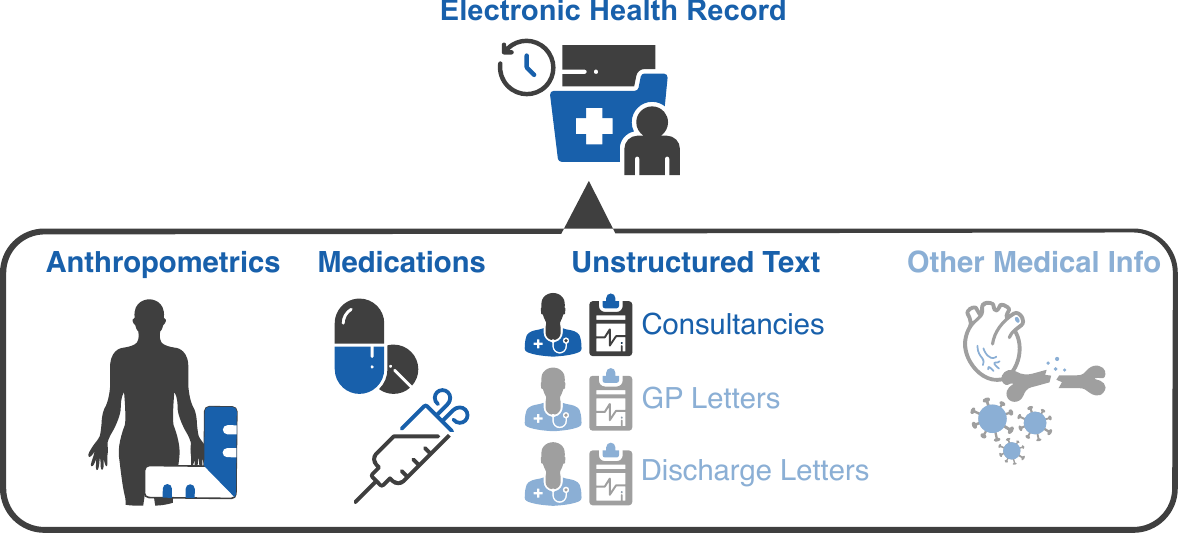}
    \caption{Graphical representation of an EHR. Highlighted items are the one used in this study, the not highlighted one are present but discarded.}
    \label{fig:dataset_format}
\end{figure}

\subsection{Consult texts}\label{ss:consults}
In a geriatric outpatient clinic, a consultation report (consultverslag) serves as a structured medical document summarizing the specialist’s assessment and recommendations following a patient visit. These reports include a multidimensional evaluation, addressing the patient’s primary complaint, comorbidities, cognitive status, functional ability, and psychosocial factors. 

The document follows a standardized format, beginning with the specialty and consultation type, followed by patient demographics, medical history, and presenting symptoms. A clinical examination includes cognitive screening (e.g., MMSE), fall risk assessment, and relevant diagnostic findings. The conclusion section presents a differential diagnosis, considering age-related conditions such as dementia, Parkinson disease, or polypharmacy-related effects. 
The report outlines a treatment plan, including referrals to other specialists, medication adjustments, and recommendations for follow-up care. The structured format ensures continuity of care and facilitates interdisciplinary collaboration in managing the complex needs of older adults.
In our dataset, we have multiple consultation reports per patient, each associated with a specific date allowing for longitudinal tracking of patient progress and adjustments in care.
\subsection{Anthropometrics}
The \acs{ehr} contains a variety of anthropometric data, including age, gender, and potentially other measurements such as BMI. However, due to the unstructured nature of the \acs{ehr}, as highlighted in the literature, extracting and organizing these data points into a usable format presents significant challenges. In our dataset, we focused on using age and gender because these variables are typically well-structured and easier to extract. Other anthropometric measures, such as BMI, were not consistently available or structured in a way that would allow for reliable inclusion in the analysis. The difficulties in processing and polishing EHR data to a usable state limit the integration of more detailed anthropometric information, despite its potential value in patient assessment.

\subsection{Medication Processing}
Medications represent a critical unstructured feature, originally recorded in the EHRs by their trade or generic names. To ensure semantic consistency and resolve variability between different denominations for the same active ingredient, all medication entries were mapped to the Anatomical Therapeutic Chemical (ATC) classification system~\cite{atc}, developed by the World Health Organization.
For each patient, the pharmacological history spanning the observation year was transformed from raw prescription names into a list of decompressed ATC code descriptions. This standardization step converts heterogeneous drug data into a unified textual format, serving as the input for the semantic embedding and aggregation pipeline detailed in the Methodology.

\section{Methodology}
\begin{figure}[H]
    \centering
    \includegraphics[width=\textwidth]{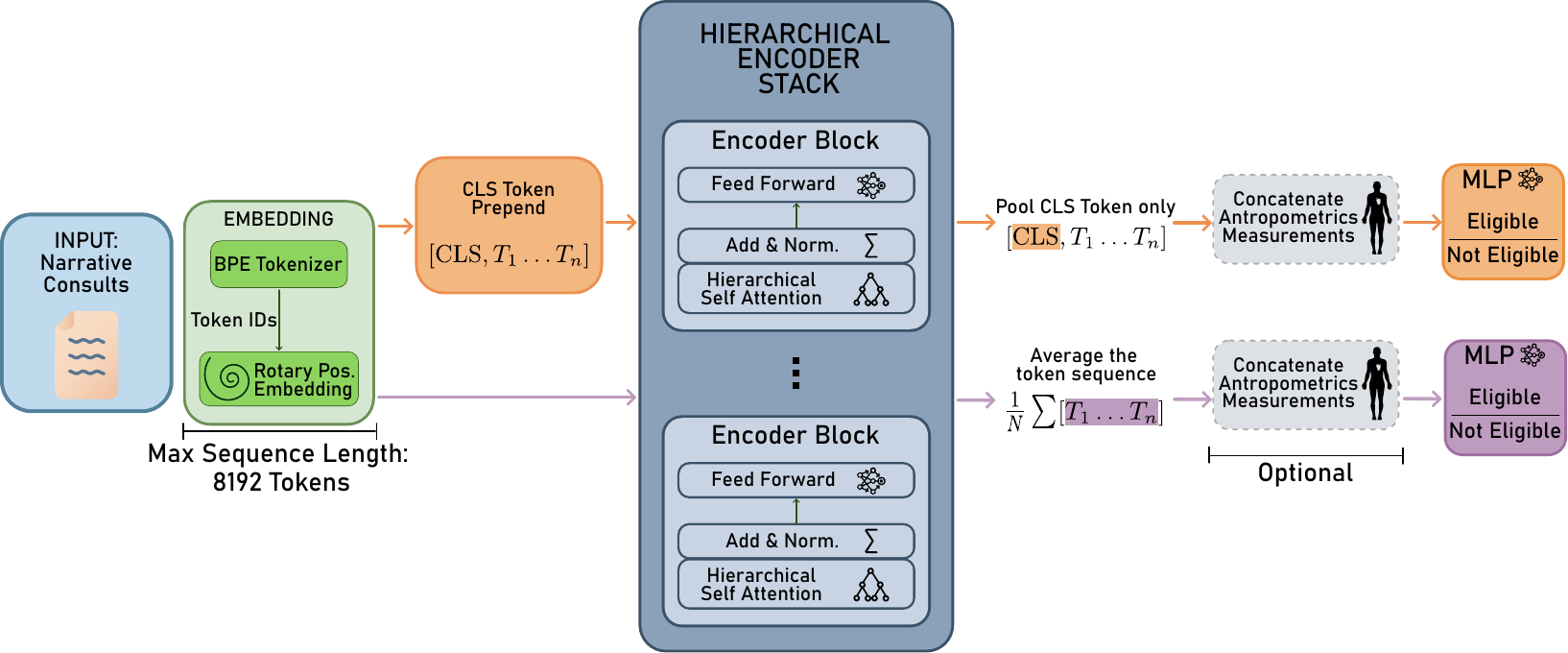}
    \caption{Schematic overview of the Hierarchical Transformer classification pipeline. The architecture processes concatenated consults via BPE tokenization and hierarchical encoding, utilizing CLS Classification Token (orange arrows) or Global Average Pooling (purple arrows) strategies before late fusion with anthropometric data (or not).}
    \label{fig:htransformer_cls}
\end{figure}

Data from consultations, medications, and anthropometric measurements acquired for a single year were selected from patients' EHRs, utilizing the most recent medical appointment as the reference time-point. To systematically evaluate the impact of integrating these distinct modalities across all our evaluated models, we investigated two modeling strategies: a unimodal \textit{text-only} approach and a multimodal \textit{late fusion} approach (Figure~\ref{fig:htransformer_cls})~\cite{Wang2022-hv,Rajkomar2018-hj}. The text-only strategy serves as a baseline, relying solely on the text encoders to classify the clinical narratives. Conversely, the late fusion strategy processes the unstructured text independently and concatenates the resulting feature representation with the structured data vectors (medications and anthropometrics) just prior to the final classification head. We selected this specific fusion architecture because it allows the core sequence-modeling networks to be utilized without requiring complex internal modifications for multimodal processing, while still empirically maintaining robust predictive performance~\cite{ngiam2011multimodal}.

To operationalize this late fusion strategy, the supplementary modalities required mathematical representation. To incorporate the pharmacological data, we leveraged the Anatomical Therapeutic Chemical (ATC) codes already recorded in the patient EHRs~\cite{atc}. We extracted the decompressed textual description associated with each code to capture the semantic nuance of the prescribed treatments. Subsequently, these descriptive strings were converted into numerical embeddings using the BioLORD-2023 model, a Sentence-Transformer~\cite{BioLORD,10.1093/bioinformatics/btz682}. This model was selected for its pre-training on the UMLS medical ontology, which includes descriptions of ATC codes, enabling it to map semantically similar drugs to proximate coordinates within the vector space~\cite{Beam2020-zx}. By feeding the ATC descriptions into this model, we obtained 768-dimensional embeddings, which were subsequently condensed into a single representative vector per patient to serve as the structured input for the late fusion layer.

We benchmarked our proposed architecture against three distinct families of modeling paradigms, applying both the text-only and late fusion strategies where applicable. First, to establish traditional machine learning baselines, we employed a stratified dummy classifier and a Linear Support Vector Classifier (LinearSVC) trained on Term Frequency–Inverse Document Frequency (TF-IDF) representations~\cite{lin-etal-2023-linear,manning2008introduction}. Second, to evaluate the efficacy of attention mechanisms against convolutional approaches for sequential medical text, we implemented a custom one-dimensional ResNet~\cite{He2015}. Finally, we compared our supervised architectures against general-purpose \acs{llm}s applied in a zero-shot classification setting~\cite{zhu2024promptinglargelanguagemodels,labrak-etal-2024-zero}.

To process the longitudinal consultation narratives, we developed a custom encoder-only Transformer architecture designed specifically for large-context sequences. The input text is first encoded using a pre-trained tokenizer, mapping each token ID to a 512-dimensional embedding space. The sequence is padded or truncated to a maximum length of 8192 tokens. To track token positions within the input text, a rotary positional embedding~\cite{rope} is applied. For the final classification mechanism, we investigated two distinct aggregation strategies. The first approach utilizes a Classification (CLS) token~\cite{Devlin2019} prepended to the embedded sequence. During the back-propagation phase, this randomly initialized token inflates with global classification information; after the Transformer Encoder stack, only the CLS token is pooled and passed to a Multi-Layer Perceptron (MLP) to output class predictions, as detailed in Fig.~\ref{fig:htransformer_cls}. Alternatively, we explored a Global Average Pooling strategy, where the output token sequence is simply averaged to form the final representation, eliminating the need for a CLS token.
Each Transformer Encoder layer is composed of a Hierarchical Attention Layer and a Feed-Forward network layer, every output of those layer is added via skip connection and layer-normalised as explained in \cite{Vaswani2017}. The token sequence flows into the Transformer encoder where hierarchical attention is computed as shown in~\cite{ZhuSoricut2021}. This approach allows large context-length text to be processed with acceptable computational time and memory complexity. This modified version of attention requires that the input is padded to a power of two, as attention is computed hierarchically on token pairs, prioritizing local dependencies over distant ones~\cite{ZhuSoricut2021}. In parallel, we implemented a custom ResNet-1D architecture tailored to sequential representations, providing a convolutional baseline for comparison with Transformer-based methods.

The hyperparameters and structural configurations detailed in Table~\ref{tab:model_summary} were selected following a grid search constrained by our available GPU resources. We observed that variations among the top-performing configurations yielded no statistically significant improvements; therefore, for clarity and brevity, we report only the hyperparameter set that achieved the highest validation metrics.

\begin{table}[ht]
    \centering
    \caption{\textbf{Model Architecture and Hyperparameter Summary}. detailed breakdown of the Hierarchical Transformers (HTrans), ResNet-1D, and LinearSVC baseline. Note the distinct classification head structures and dropout rates.}
    \label{tab:model_summary}
    \resizebox{\textwidth}{!}{
    \begin{tabular}{lcccc}
        \toprule
        \textbf{Feature} & \textbf{HTrans (Avg)} & \textbf{HTrans (CLS)} & \textbf{ResNet} & \textbf{LinearSVC} \\
        \midrule
        \textit{Architecture Base} & \multicolumn{2}{c}{Hierarchical Transformer} & ResConsultNet (CNN) & SVM (Linear) \\
        \textit{Input Representation} & \multicolumn{2}{c}{Learned Embeddings} & Learned Embeddings & TF-IDF \\
        \textit{Aggregation} & Global Avg Pool & CLS Token & Max Pooling (Inter-layer) & Hyperplane Dist. \\
        \midrule
        \textbf{Structural Params} & & & & \\
        Total Layers & $3$ & $3$ & $8$ & -- \\
        Block Composition & \multicolumn{2}{c}{PreNorm $\to$ Attn/FF $\to$ Add} & Conv1D $\to$ BN $\to$ ReLU & -- \\
        Positional Embedding & \multicolumn{2}{c}{Rotary (RoPE)} & -- & -- \\
        Heads / Filters & $4$ Heads & $4$ Heads & $16$ (Base) & -- \\
        Dim Head / Kernel & $32$ & $32$ & $3$ & -- \\
        Block Size & $32$ & $32$ & -- & -- \\
        FF Multiplier & $4$ & $4$ & -- & -- \\
        \midrule
        \textbf{Classification Head} & & & & \\
        Structure (Hidden Dims) & \multicolumn{2}{c}{$256 \to 128$} & $20$ & -- \\
        Dropout (Head) & \multicolumn{2}{c}{$0.2$} & $0.5$ & -- \\
        Normalization (Head) & \multicolumn{2}{c}{BatchNorm1d} & BatchNorm1d & -- \\
        \midrule
        \textbf{LinearSVC Params} & & & & \\
        Regularization ($C$) & -- & -- & -- & $1.0$ \\
        Loss Function & -- & -- & -- & Squared Hinge \\
        Tolerance & -- & -- & -- & $1 \times 10^{-4}$ \\
        \bottomrule
    \end{tabular}
    }
\end{table}

\subsection{Training and Optimization}
All deep learning models were trained using the AMSGrad~\cite{amsgrad} optimizer. Given the unbalanced nature of the dataset, we employed a Stratified 5-fold cross-validation strategy. This ensures that the proportion of samples for each class is preserved in each fold, providing a robust evaluation metric. The specific training hyperparameters are detailed in Table~\ref{tab:training_config}.

\begin{table}[H]
    \centering
    \caption{\textbf{Training \& Optimization Configuration}. These hyperparameters are shared across all model variations in the experimental setup.}
    \label{tab:training_config}
    \begin{tabular}{ll}
        \toprule
        \textbf{Parameter} & \textbf{Value} \\
        \midrule
        Epochs & $30$ \\
        Batch Size & $12$ \\
        Learning Rate & $3 \times 10^{-5}$ \\
        Weight Decay & $1 \times 10^{-4}$ \\
        Optimizer Flag & Adam (AMSGrad) \\
        \midrule
        \textit{Data Split Strategy} & \\
        Cross-Validation & Stratified K-Fold \\
        K-Fold Splits & $5$ \\
        Test Size & $0.2$ ($20\%$) \\
        Seed & $42$ \\
        \bottomrule
    \end{tabular}
\end{table}
Finally, we evaluated few \acs{llm}s in a zero-shot classification setting with a two-prompt approach with the first prompt for translation from Dutch to English and the second prompt to extract the binary label, where we add an extract of the CVRM guidelines to the system prompt. The \acs{llm}s used were GPT-4o, GPT-4o-mini, GPT-4.1 and GPT-4.1-mini\footnote{The O-models and the GPT-5.1 models were not available yet in the secure Azure environment for West-Europe.}. Before the API is called the consult texts are de-identified using DEDUCE~\cite{Menger2018}\fref{https://github.com/vmenger/deduce}. We prepend the consult texts with the age and the gender of the patient.

Note that the CVRM guidelines are over 400 pages. In the spirit of this approach we prompted GPT 5.2 to make an
extractive summary of the guidelines regarding the core CVRM factors, see appendix \ref{app:cvrm_summary}.
We append this extractive summary to the system prompt.

\begin{promptgroupenv}[Two-step setup]
\label{group:multi_prompt}
    \begin{prompt}[System]\label{prompt:pr1_multi}
\begin{prompttext}
    You are a faithful and truthful label extractor in the cardio/geriatrics domain. You are an expert in cardiovascular risk management. You assign people to the cardiovascular risk management regime based on the Dutch guidelines. This is an extract of the CVRM guidelines: [SUMMARY OF CVRM GUIDELINES]
\end{prompttext}
    \end{prompt}
    \begin{prompt}[Translation]\label{prompt:pr2_multi}
\begin{prompttext}
    Translate this Dutch geriatrics consult to English.
\end{prompttext}
    \end{prompt}
    \begin{prompt}[Extraction]\label{prompt:pr3_multi}
\begin{prompttext}
We want to know whether the patient, based on this medical consult text and the CVRM guidelines, has an elevated risk for cardiovascular disease. Only respond with yes / no.
\end{prompttext}
    \end{prompt}
\promptgroupcaption{Two-step prompt, first we translate, then we extract.}
\end{promptgroupenv}

\section{Results and Discussion}
Table~\ref{tab:ml_performance} reports the performance of all models using consult texts only as input, while the late-fusion approach, which combines textual data with patient anthropometric features, is summarized in Table~\ref{tab:ml_performance_all}.
Both tables report F1-score, Precision, Recall, and Matthews Correlation Coefficient (MCC) as evaluation metrics (reported below).
\begin{equation}
    \begin{aligned}
        \text{Precision} &= \frac{TP}{TP + FP} \\[1ex]
        \text{Recall} &= \frac{TP}{TP + FN} \\[1ex]
        \text{F1-score} &= 2 \times \frac{\text{Precision} \times \text{Recall}}{\text{Precision} + \text{Recall}} = \frac{2TP}{2TP + FP + FN} \\[1ex]
        \text{MCC} &= \frac{TP \times TN - FP \times FN}{\sqrt{(TP + FP)(TP + FN)(TN + FP)(TN + FN)}}
    \end{aligned}
    \label{eq:performance_metrics}
\end{equation}

\begin{table}[H]
    \centering
    \caption{Performance of \ac{cvrm} prediction based on only consult texts as input. Values are shown as mean (std).}
    \label{tab:ml_performance}
    \begin{threeparttable}
    \begin{tabular}{lcccc}
         Models &  \bf F1-Score (\%) &  \bf Precision (\%)& \bf Recall (\%)& \bf MCC-Score$^{\oplus}$\\
         \hline
  Dummy Classifier            & 31.17 (0.53) & 19.36 (0.33) & 79.94 (1.47)& -0.0013 (0.016)\\
  Linear SVC$^\triangle$      & 86.13 (0.59) & 82.80 (0.57) & 92.78 (0.48) & 0.749 (0.011)\\
  \bf H-transformer 1D$^{\dag}$   & \bf 89.73 (3.39) &  \bf 89.73 (3.39) &  \bf 91.48 (1.25)& \bf 0.718 (0.049)\\
  \bf H-transformer 1D$^{\ddag}$  & \bf 91.48 (1.25) &  \bf 91.48 (1.25) &   \bf 89.73 (3.39)& \bf 0.646 (0.144)\\
  ResNet1D                    & 78.54 (2.16) & 78.54 (2.16) & 78.54 (2.16) & 0.371 (0.102)\\
  Zero-Shot LLM  GPT-4o-mini  & 34.81 & 21.31 & 94.8 & 0.12\\
  Zero-Shot LLM  GPT-4o       & 35.29 & 21.63 & 95.7 & 0.14\\
  Zero-Shot LLM  GPT-4.1-mini & 34.96 & 21.7 & 89.3 & 0.12 \\
  Zero-Shot LLM  GPT-4.1      & 33.45 & 20.3  & 93.3  & 0.07 \\
  \hline
    \end{tabular}
    \begin{tablenotes}[flushleft]
        \footnotesize
        \item $\oplus$: Matthew's Correlation Coefficient; $\dag$: Hierarchical Transformer with Classification Token Pooling; $\ddag$: Hierarchical Transformer with average pooling; $\triangle$: Support Vector Classifier with Linear Kernel;
    \end{tablenotes}
    \end{threeparttable}
\end{table}
As shown in Table~\ref{tab:ml_performance} and Table~\ref{tab:ml_performance_all}, hierarchical Transformer models consistently outperform traditional machine learning baselines and state-of-the-art zero-shot \acs{llm}s in both text-only and late-fusion settings, demonstrating that their architecture effectively captures the complex structure of clinical texts. 
In the text-only setting (Table~\ref{tab:ml_performance}), the classification token variant ($^{\dag}$) achieved an F1-score of 92.48\% with an MCC of 0.758, while the average pooling variant ($^{\ddag}$) yielded slightly lower yet still strong performance (F1-score: 91.02\%, MCC: 0.730). When anthropometric features were incorporated via late fusion (Table~\ref{tab:ml_performance_all}), both variants maintained top performance, with F1-scores ranging from 89.73\% to 91.48\% and MCC values between 0.646 and 0.718.

\begin{table}[H]
    \centering
    \caption{Performance of \ac{cvrm} prediction based on consult texts and anthropometric measures as input. Values are shown as mean (std).}
    \label{tab:ml_performance_all}
    \begin{threeparttable}
    \begin{tabular}{lcccc}
         \textbf{Models} &  \textbf{F1 (\%)} &  \textbf{Precision (\%)} & \textbf{Recall (\%)} & \textbf{MCC}$^{\oplus}$ \\
         \hline
         Dummy Classifier & 50.76 (0.0) & 50.75 (0.0) & 50.78 (0.0) & 0.015 (0.0) \\
         Linear SVM$^{\triangle}$ & 86.30 (0.56) & 82.97 (0.54) & 92.87 (0.43) & 0.752 (0.010) \\
         \bf H-transformer1D$^{\dag}$ & \textbf{92.48 (0.67)} & \textbf{92.48 (0.67)} & \textbf{92.48 (0.67)} & \textbf{0.758 (0.024)} \\
         \bf H-transformer1D$^{\ddag}$ & \bf 91.02 (1.11) & \bf 91.02 (1.11) & \bf 91.02 (1.11) & \bf 0.730 (0.028) \\
         ResNet1D & 85.48 (0.69) & 85.48 (0.69) & 85.48 (0.69) & 0.565 (0.023) \\
         \hline
    \end{tabular}
    \begin{tablenotes}[flushleft]
        \footnotesize
        \item $\oplus$: Matthews Correlation Coefficient; $\dag$: Hierarchical Transformer with Classification Token Pooling; $\ddag$: Hierarchical Transformer with Average Pooling; $\triangle$: Support Vector Classifier with Linear Kernel;
    \end{tablenotes}
    \end{threeparttable}
\end{table}

Although the choice of pooling strategy has a minimal impact on overall performance, the classification token-based variant consistently achieves the highest scores, suggesting that this strategy is better suited to capturing the most informative content present in medical consultation texts.
Among the evaluated baselines, the \acs{svm} achieved strong performance across the considered metrics. Despite its relatively simple architecture, \acs{svm} yielded results that were only moderately lower than those obtained by our proposed model, confirming its effectiveness in this clinical prediction task.
This result is consistent with previous evidence highlighting the robustness of \acs{svm} in clinical and healthcare applications, particularly in settings characterized by limited data availability and high-dimensional feature spaces. 
Prior to the widespread adoption of transformer-based architectures, \acs{svm}s were considered a strong methodological standard in clinical machine learning and continue to represent a competitive baseline in such scenarios~\cite{G2025}.
In contrast, ResNet shows substantially lower performance, particularly when trained using only medical consultations. 
This suggests that deep learning architectures may struggle in clinical tasks when training data are limited. 
Such models typically rely on large amounts of labeled data to effectively learn task-relevant representations, a requirement that is often difficult to satisfy in real-world clinical settings.
Indeed, this trend is even more pronounced for \acs{llm}s evaluated in a zero-shot configuration, which exhibited the weakest performance. 
These findings suggest that, in the absence of domain adaptation or fine-tuning, \acs{llm}s may lack adequate understanding of the clinical context required for reliable prediction.
Nonetheless, all models achieved performance levels superior to the dummy classifier, demonstrating their capacity to extract meaningful information from both structured and unstructured data.
All in all, the superior performance of the H-transformer-1D highlights the importance of hierarchical attention when processing the complex, multi-dimensional evaluations typical of geriatric consultation reports. 
Unlike standard models, the hierarchical approach efficiently captures long-range dependencies within the clinical history, which is essential given that geriatric patients often present with multiple comorbidities and polypharmacy that synergistically contribute to cardiovascular risk.

\section{Conclusion}
This study demonstrates that deep learning architectures can successfully automate the identification of cardiovascular risk management eligibility using unstructured electronic health records. The proposed custom Hierarchical Transformer effectively handled longitudinal clinical narratives, yielding higher classification performance compared to both traditional machine learning baselines and general-purpose Large Language Models (LLMs) applied in a zero-shot setting. 

Our findings indicate that while general-purpose generative models offer broad capabilities, they currently lack the precision required for this specific non-English clinical task without explicit supervision. Although evaluating fine-tuned, domain-specific clinical LLMs presents an interesting future avenue, the choice to develop a bespoke encoder architecture was driven by a clear operational rationale. Strict data privacy regulations necessitate that sensitive patient EHRs remain within secure, compute-constrained, on-premise hospital environments. In such infrastructural settings, fine-tuning massive autoregressive generative models is often computationally prohibitive. Conversely, our proposed Transformer, while architecturally sophisticated in its attention mechanism, maintains a significantly reduced training footprint. It is inherently optimized for sequence classification rather than text generation, making it both highly efficient to train from scratch and practically deployable. Consequently, task-specific encoder architectures remain a necessary and pragmatic solution for reliable risk stratification.

By mitigating the selection errors associated with manual administrative coding, this automated approach supports the operational goals of a Learning Healthcare System. Future implementation of such models in routine practice offers a scalable, privacy-compliant method to utilize accumulated patient data for continuous quality improvement and more accurate geriatric risk assessment.

\section*{Declarations}

\subsection*{Ethics approval and consent to participate}
The \ac{umcu} quality assurance research officer confirmed under project number 22U-0292 that this study does not fall under the scope of the Dutch Medical Research Involving Human Subjects Act (WMO) and therefore does not require approval from an accredited medical ethics committee. The study was performed compliant with local legislation and regulations. All patient data were deidentified in compliance with the European Union General Data Protection Regulation, and as a result, written informed consent was not required by the \ac{umcu} ethical committee.

\subsection*{Availability of data and materials}
The datasets generated and/or analysed during the current study are not publicly available due to potential privacy-sensitive information, but are available from the corresponding author upon reasonable request and local institutional approval. Research code is publicly available on GitHub, via \href{/link/to/coderepository}{the code repository for this work}.

\subsection*{Competing interests}
The authors declare that they have no competing interests.

\subsection*{Funding}
The work was supported in part by the ODIN Project, as part of the European Union’s Horizon 2020 Research and Innovation Program, under Grant 101017331.

\subsection*{Authors' contributions}
\textbf{Jacopo Vitale}: Conceptualisation, Methodology, Data Curation, Writing - Original Draft, Investigation. \textbf{David Della Morte}: Methodology, Investigation, Data Curation, Writing original draft. \textbf{Luca Bacco}: Methodology, Writing Review \& Editing. \textbf{Mario Merone}: Writing Review \& Editing, Supervision. \textbf{Mark C.H. de Groot}: Writing Review \& Editing, Resources. \textbf{Saskia Haitjema}: Resources, Writing Review \& Editing. \textbf{Leandro Pecchia}: Project Administration, Funding Acquisition, Supervision. \textbf{Bram Van Es}: Conceptualisation, Writing Original Draft, Data Curation, Methodology, Investigation, Supervision.

\appendix 
\newpage
\section{GPT summary of CVRM guidelines}
\label{app:cvrm_summary}
\textit{Prompt:}
\begin{verbatim}
I want to use ChatGPT to extract labels from patient consults regarding cardiovascular 
risk management. To guide the model I want to include guidelines but these are expansive,
with over 400 pages of text. Could you extract from these guidelines the core points 
regarding co-morbidities, symptoms, prior history, risk scores and other factors that 
are likely included in these consults? 
Keep it brief. No extra explanatory text, just an extractive summary.
\end{verbatim}
\textit{Response:}
\begin{verbatim}
## Core CVRM factors commonly documented in patient consults

### Demographics
- Age  
- Sex  

### Symptoms / Clinical presentation
- Chest pain / angina  
- Dyspnea  
- Palpitations (e.g. atrial fibrillation)  
- Neurological deficits / TIA / stroke symptoms  
- Claudication (peripheral arterial disease)  
- Signs of heart failure  

### Prior history (established disease)
- Documented atherosclerotic cardiovascular disease  
  (coronary artery disease, myocardial infarction, stroke/TIA, 
  peripheral arterial disease, aortic aneurysm)
- Diabetes mellitus (type 1 or 2; duration, complications)
- Chronic kidney disease (eGFR, albuminuria)
- Familial hypercholesterolemia
- Atrial fibrillation
- Heart failure
- Hypertension
- Prior revascularization or vascular procedures

### Risk factors (anamnestic)
- Smoking status (current/former, pack-years)
- Family history of premature cardiovascular disease
- Diet quality
- Physical inactivity / sedentary behavior
- Alcohol use
- Psychosocial stress, depression, low socioeconomic status

### Risk factors (clinical / measurements)
- Systolic blood pressure
- Body mass index (BMI)
- Waist circumference

### Laboratory factors
- LDL-cholesterol
- Non-HDL-cholesterol
- Total cholesterol
- HDL-cholesterol
- Triglycerides
- Fasting glucose
- Serum creatinine / eGFR
- Urine albumin–creatinine ratio

### Risk scores
- SCORE2 (ages 40–70)
- SCORE2-OP (ages 70–90)
- SMART2 / SMART-REACH (established cardiovascular disease)
- DIAL2 (diabetes mellitus)

### Risk modifiers / additional factors
- Coronary artery calcium score
- Psychosocial factors
- Ethnic background
- Chronic inflammatory diseases  
  (rheumatoid arthritis, psoriatic arthritis, ankylosing spondylitis)
- COPD
- Gout
- HIV infection
- Inflammatory bowel disease
- Obstructive sleep apnea
- History of pre-eclampsia or pregnancy-related hypertension
- Severe psychiatric disorders
- Prior chemo- or radiotherapy

### Treatment-related context often noted
- Current antihypertensive therapy
- Current lipid-lowering therapy (statins, ezetimibe, PCSK9 inhibitors)
- Blood pressure target attainment
- LDL-C target attainment
- Medication adherence
- Polypharmacy and frailty (especially in older adults)

\end{verbatim}

\bibliography{bibliography} 

\end{document}